\documentclass[11pt, letterpaper]{arxiv}

\usepackage[utf8]{inputenc} 
\usepackage[T1]{fontenc}
\usepackage{csquotes}
\usepackage{url}
\usepackage{microtype}      

\usepackage[dvipsnames, table]{xcolor}

\definecolor{ysdarkpurple}{HTML}{4E2399}
\definecolor{ysshallowpurple}{HTML}{E6DBFF}
\definecolor{ysdarkred}{HTML}{8c2824}
\definecolor{ysshallowred}{HTML}{F8D7D7}
\definecolor{ysdarkblue}{HTML}{005E99}
\definecolor{ysshallowblue}{HTML}{CCEBFF}
\definecolor{ysdarkgrey}{HTML}{333333}
\definecolor{ysshallowgrey}{HTML}{E5E5E5}
\definecolor{lightblue}{rgb}{0.22,0.45,0.70}

\definecolor{ColorGrok}{HTML}{FFFDE7}
\definecolor{ColorPplx}{HTML}{EFFDFE}
\definecolor{ColorOpenAI}{HTML}{F2F2F2}
\definecolor{ColorGemini}{HTML}{E6F4FE}
\definecolor{ColorClaude}{HTML}{FFF3EB}
\definecolor{SectionHeaderColor}{HTML}{FFFFFF}
\colorlet{DarkerColorClaude}{ColorClaude!95!black}
\colorlet{DarkerColorPplx}{ColorPplx!95!black}
\colorlet{DarkerColorGemini}{ColorGemini!95!black}
\colorlet{DarkerColorOpenAI}{ColorOpenAI!95!black}
\colorlet{DarkerColorGrok}{ColorGrok!95!black}

\definecolor{rliableolive}{HTML}{BBCC33}
\definecolor{rliableblue}{HTML}{77AADD}
\definecolor{rliablered}{HTML}{EE8866}
\definecolor{editInitialResponse}{RGB}{255, 235, 156}
\definecolor{editBacktrack}{RGB}{0, 0, 139}
\definecolor{editRevisedResponse}{RGB}{255, 182, 193}
\definecolor{highlightmistake}{RGB}{255, 179, 179} 
\definecolor{highlightcorrect}{RGB}{179, 255, 179} 

\definecolor{codegray}{gray}{0.9}
\definecolor{codepurple}{rgb}{0.58,0,0.82}
\definecolor{codeblue}{rgb}{0.25,0.5,0.5}

\usepackage{amsmath}
\usepackage{amssymb}
\usepackage{mathtools}
\usepackage{amsthm}
\usepackage{amsfonts}
\usepackage{dsfont}
\usepackage{nicefrac}
\usepackage{pifont}         
\usepackage{etoolbox}       

\usepackage{amsmath,amsfonts,bm}

\def\eqref#1{Eq.~\ref{#1}}

\def\1{\bm{1}}

\DeclareMathAlphabet{\mathsfit}{\encodingdefault}{\sfdefault}{m}{sl}
\SetMathAlphabet{\mathsfit}{bold}{\encodingdefault}{\sfdefault}{bx}{n}

\usepackage{graphicx}
\usepackage[inkscapelatex=false]{svg}
\usepackage{wrapfig}
\usepackage{caption}
\usepackage{subcaption}
\usepackage{enumitem}       

\usepackage{booktabs}       
\usepackage{multirow}
\usepackage{multicol}
\usepackage{tabularx}
\usepackage{array}
\usepackage{makecell}
\usepackage{arydshln}       

\newcolumntype{C}[1]{>{\centering\arraybackslash}p{#1}}
\newcolumntype{Y}{>{\centering\arraybackslash}X}

\usepackage{algorithm}
\usepackage{algpseudocode}
\usepackage{listings}
\usepackage{fancyvrb}
\usepackage{fvextra}
\usepackage[most]{tcolorbox}
\tcbuselibrary{breakable, skins, listings}

\DefineVerbatimEnvironment{Verbatim}{Verbatim}{breaklines=true, breakanywhere=true}

\lstdefinelanguage{YAML}{
  morekeywords={selector, sequence, condition, task, no},
  keywordstyle=\color{codeblue}\bfseries,
  ndkeywords={},
  sensitive=false,
  comment=[l]{\#},
  morecomment=[s]{/*}{*/},
  commentstyle=\color{dkgreen}\ttfamily,
  string=[b]",
  stringstyle=\color{codepurple}\ttfamily,
  morestring=[b]',
  morestring=[b]`,
  identifierstyle=\ttfamily,
  backgroundcolor=\color{codegray},
  basicstyle=\ttfamily\footnotesize,
  breaklines=true,
  captionpos=b,
  frame=single,
  numbers=left,
  numberstyle=\tiny\color{gray},
  numbersep=5pt,
  tabsize=2,
  showspaces=false,
  showstringspaces=false,
  showtabs=false,
  xleftmargin=1em,
}

\newtcbinputlisting{\promptbox}[2][]{
  enhanced, breakable, colback=ysshallowblue, colframe=ysdarkblue,
  fonttitle=\bfseries, title=#2, listing only,
  listing options={
    language=YAML, basicstyle=\ttfamily\small, breaklines=true,
    breakatwhitespace=true, postbreak=\mbox{\textcolor{red}{$\hookrightarrow$}\space},
    showstringspaces=false, upquote=true,
  }, #1
}

\newtcolorbox{analysisbox}[1][]{
    enhanced jigsaw, colback=white, colframe=blue!75!black,
    fonttitle=\bfseries, boxsep=5pt, left=5pt, right=5pt, top=5pt, bottom=5pt,
    title=#1,
}

\newtcolorbox{solutionbox}{
  colframe=black, colback=gray!10, boxrule=1pt, arc=0pt,
  title=, fonttitle=\bfseries
}
\newenvironment{sol}{\begin{solutionbox}}{\end{solutionbox}}
\newcommand{\BeginSol}{\begin{sol}}
\newcommand{\EndSol}{\end{sol}}

\AtBeginEnvironment{multline}{
    \setlength{\abovedisplayskip}{2pt}
    \setlength{\belowdisplayskip}{2pt}
    \setlength{\abovedisplayshortskip}{1pt}
    \setlength{\belowdisplayshortskip}{1pt}
}
\theoremstyle{plain}

\theoremstyle{definition}

\theoremstyle{remark}

\setboolean{logo}{False}
\usepackage[comma,numbers,sort,compress]{natbib}
\usepackage[colorlinks=true, allcolors=blue]{hyperref}
\usepackage[all]{hypcap}
\usepackage{crossreftools} 

\hypersetup{
    colorlinks = true,
    citecolor = {magenta},
}

\usepackage[capitalize,noabbrev]{cleveref}

\newcommand{\refsec}[1]{\S\ref{#1}}
\newcommand{\ours}{WiA-LLM}

\title{What-If Analysis of LLMs: Explore the Game World Using Proactive Thinking}

\author[1]{Yuan Sui}
\author[2]{Yanming Zhang}
\author[3]{Yi Liao}
\author[3]{Yu Gu}
\author[3]{Guohua Tang}
\author[3]{Zhongqian Sun}
\author[3]{\protect\\Wei Yang}
\author[1]{Bryan Hooi}

\affil[1]{NUS}
\affil[2]{ZJU}
\affil[3]{Tencent}
\correspondingauthor{yuan.sui@u.nus.edu}

\begin{document}
\maketitle

\textbf{Abstract:}
LLMs struggle with decision-making in high-stakes environments like MOBA games, primarily due to a lack of proactive reasoning and limited understanding of complex game dynamics. To address this, we propose What-if Analysis LLM (WiA-LLM), a framework that trains an LLM as an explicit, language-based world model. Instead of representing the environment in latent vectors, WiA-LLM uses natural language to simulate how the game state evolves over time in response to candidate actions, and provides textual justifications for these predicted outcomes. 
WiA-LLM is trained in two stages: supervised fine-tuning on human-like reasoning traces, followed by reinforcement learning with outcome-based rewards based on the alignment between predicted and actual future states. In the Honor of Kings (HoK) environment, WiA-LLM attains 74.2\% accuracy (27\%$\uparrow$ vs. base model) in forecasting game-state changes. In addition, WiA-LLM demonstrate strategic behavior more closely aligned with expert players than purely reactive LLMs, indicating enhanced foresight and expert-like decision-making.

\section{Introduction}
\label{sec:introduction}

What-If Analysis (WIA), as the name suggests, is a systematic decision-making approach that addresses hypothetical questions such as \enquote{\textit{What if we take this action? How will it affect the final outcome?}}. WIA enables decision-makers to simulate hypothetical scenarios by altering specific action variables and assessing their potential implications~\cite{what-if-analysis-business}. This methodology is valuable for strategic planning, risk assessment, and enhancing the explainability of the decision-making process~\cite{tang2025timeseries}.

Despite its potential, the integration of WIA capabilities into large language models (LLMs) remains underexplored. This reveals a critical limitation of current LLMs when applied to dynamic, high-stakes scenarios such as strategic planning, risk assessment, and real-time decision making. While existing LLMs excel at \textit{reactive thinking}~\cite{liao2025think,cot_23,li2025codei0o0}-generating answers based on the current context and their prior knowledge-they lack mechanisms for \textit{proactive thinking}, which is essential for forecasting the consequences of actions before they occur. This limitation is particularly pronounced in dynamic environments, where each action may trigger a series of cascading effects~\cite{hok_env_2022, hu2024pokellmonhumanparityagentpokemon,meta_reasoner_2025}. Understanding the consequences of different actions not only clarifies the environment but also provides deeper intuition for making informed decisions.

To address this gap, we introduce \textbf{\ours{}}, a novel framework that endows LLMs with proactive thinking capabilities through explicit world modeling. By leveraging environmental feedback via reinforcement learning (RL), \ours{} learns to forecast the outcomes of different actions on the entire game state. Our core insight draws from human cognition: \textbf{``Look before you leap''}, i.e., one should consider possible consequences or dangers before acting. We formalize this process as explicit world modeling: $S_\Delta = f(S_t, a_t)$, where the model predicts the state transition $S_\Delta$ resulting from taking action $a_t$ in state $S_t$ using natural language. The task becomes increasingly challenging as more properties change in $S_\Delta$ (see \refsec{sec:data_collection}). Our training pipeline first applies supervised fine-tuning (SFT) on human gameplay trajectories to provide basic behavioral and environmental knowledge. It then applies RL with rule-based, verifiable rewards that compare the model's forecasts against actual environment transitions, aligning its predictions with real dynamics (see \refsec{sec:rl_mtd}). In this way, \ours{} shifts LLMs from purely reactive pattern matching to model-based forecasting, analogous to how humans mentally simulate outcomes before action.

We evaluate \ours{} in \textit{Honor of Kings} (HoK), a large-scale multiplayer online battle arena (MOBA) game~\cite{hok_env_2022}. HoK servers as an ideal testbed for three reasons. First, it exhibits \textit{high dynamic complexity}: players must adapt in real time to over one hundred heroes, shifting objectives, and coordinated team strategies. Second, it offers \textit{quantifiable states}: the game state can be encoded as JSON-structured objects with hero positions, resources, and map conditions, enabling precise and automatic reward computation. Third, it features \textit{high-stakes consequences}: a single mistimed objective contest (e.g., a dragon fight) can flip the match outcome, creating a rich space for WIA evaluation. To thoroughly test whether \ours{} can support dynamic proactive reasoning and adapt decision-making based on forecasted consequences, we construct two benchmarks in this environment (see \refsec{sec:data_collection} and \refsec{sec:experiment_setup}).

\begin{figure}[t]
  \centering
  \includegraphics[width=\linewidth]{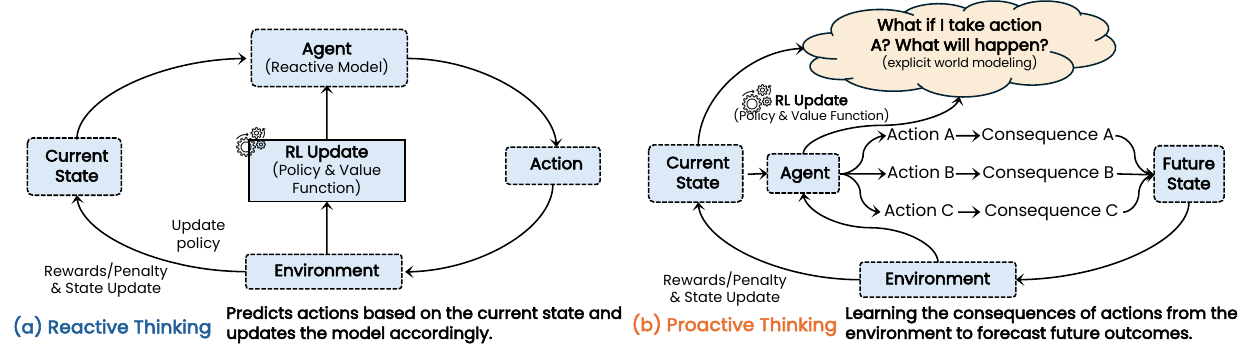}
  \caption{\textbf{Illustration of reasoning paradigms}: (a) reactive thinking, where the model selects an action given the current game state; (b) proactive thinking, where the model also forecasts the consequences of candidate actions on future game states. In this work, we focus on proactive thinking and train models to forecast the consequences of different actions.}
  \label{fig:demonstration}
\end{figure}

Our experiments demonstrate that: (1) \ours{} achieves 74.2\% forecasting accuracy on HoK scenarios, outperforming the base model (qwen3-14b) by 27\% and surpassing deepseek-r1 by 41.6\%; (2) it enhances downstream decision-making, enabling agents to exhibit strategic behavior closer to that of expert human players; (3) it consistently improves across difficulty tiers, achieving 93.9\% accuracy on the simplest forecasting tasks and 73.1\% on moderately complex tasks; (4) it demonstrates stable optimization during RL training, with reward convergence within 400 steps and no observed degradation in output quality; and (5) it maintains strong zero-shot generalization on standard academic benchmarks.

\textbf{Overall, our contributions are as follows:}
\begin{itemize}
\item We propose \ours{}, a framework with proactive thinking by forecasting the consequences of actions using environment feedback.
\item We design a training paradigm that combines SFT on human gameplay data with RL guided by rule-based, verifiable rewards, aligning model forecasts with actual environment transitions.
\item We demonstrate that \ours{} achieves strong performance in a complex MOBA environment and enables agents to exhibit strategic behavior closer to expert human players.
\end{itemize}

\section{Method: \ours{}}
\label{sec:method}

\subsection{Overview \& Problem Formulation}
Our framework, \ours{}, facilitates the transition of LLMs from reactive pattern matching to proactive world modeling. We conceptualize the What-If Analysis task as a conditional state prediction problem within a partially observable Markov Decision Process (POMDP)~\cite{wiki:Partially_observable_Markov_decision_process}. Formally, let $\mathcal{S}$ and $\mathcal{A}$ denote the game state and action space. At time step $t$, the agent observes a state $S_t$ and chooses a candidate action $a_t \in \mathcal{A}$. The objective is to approximate the transition function $\mathcal{T}(S_{t+1} | S_t, a_t)$ by predicting the future state changes $S_\Delta$, where $S_\Delta$ represents the causal impact of the action to the entire state. Unlike standard next-token prediction, this task requires the model to implicitly simulate environment dynamics and generate justifications for the state transition. 

\begin{figure*}[t]
  \centering
  \includegraphics[width=0.9\linewidth]{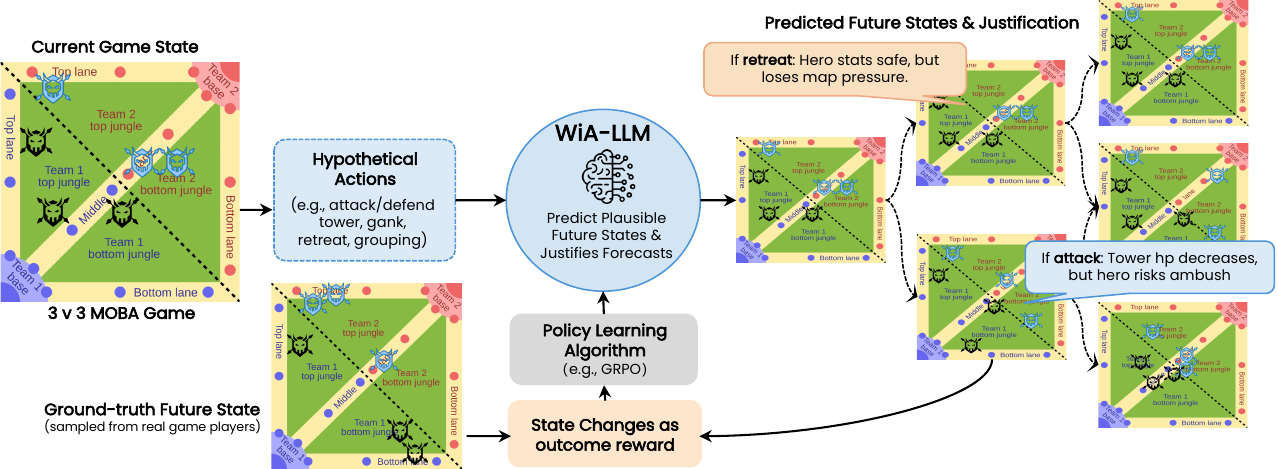}
  \caption{\textbf{Workflow} of \ours{}.
    Given the current game state and a set of hypothetical actions, the model is tasked with forecasting the potential changes to the entire game state that would result from each action, and providing justifications for these forecasts. The predicted game state changes are then compared to ground-truth values using a rule-based verifier, which is used to update the policy model. This process enables the model to perform what-if analysis by simulating action outcomes and iteratively  refining its decision-making.}
  \label{fig:grpo}
\end{figure*}

\subsection{Data Construction and Utilization}
\label{sec:data_collection}

To ground the model in realistic dynamics, we construct a scalable dataset derived from \textit{Honor of Kings (HoK)}. This dataset is pivotal for our multi-stage training pipeline, providing both the reasoning priors for SFT and the ground-truth oracle for RL verification.

\noindent\textbf{Trajectory Parsing and Difference Extraction.} We process raw gameplay logs using a strict state-parsing pipeline (Algorithm~\ref{alg:data_sampling}) to generate the base transition dataset $\mathcal{G}_D = \{(S_t, a_t, S_\Delta^*)_i\}_{i=1}^N$.
Specifically, each state $S_t$ is first serialized into a structured JSON object encompassing all visible information, such as hero attributes, turret status, and map vision, to enforce partial observability. We then parse the player's executed action $a_t$ corresponding to $S_t$ using a predefined taxonomy. To compute the target label, we calculate the ground-truth state difference $S_\Delta^*$ by comparing $S_t$ with the future state $S_{t+\delta}$ (where $\delta > 0$). This difference vector $S_\Delta^*$ captures changes in critical components $\mathcal{C} = \{\text{hero, tower, minions, dragon}\}$ and serves as the definitive label for environmental consequences.
The gathered dataset $\mathcal{G}_D$ is utilized across the two optimization stages described below. 

\noindent\textbf{Stage I: Reasoning Distillation (SFT).} To endow the model with strong reasoning capabilities, we augment the base training samples with synthetic reasoning traces. Leveraging a teacher model (DeepSeek-R1) that has access to the ground truth $(S_t, a_t, S_\Delta^*)$, we distill a "thinking process" $C_t$ that explains the causal link between the action and its outcome. This yields the SFT corpus $\mathcal{D}_{SFT} = \{(S_t, a_t, C_t, S_\Delta^*)\}$, which is used to train the model to generate structured reasoning before producing the final prediction.

\noindent\textbf{Stage II: Outcome Verification (RL)}. For the reinforcement learning stage, we revert to the original real-world data $\mathcal{G}_D$. Here, the pair $(S_t, a_t)$ serves as the prompt $q$, while the ground truth $S_\Delta^*$ is reserved as the oracle for the rule-based reward function. This ensures that the policy optimization is driven by actual environmental dynamics rather than distilled approximations.

\subsection{Policy Learning via Verifiable Rewards}
\label{sec:rl_mtd}

While SFT provides basic behavioral patterns, it lacks the ability to effectively guide the model to do self-explore against dynamic environments. We therefore employ Group Relative Policy Optimization (GRPO~\cite{grpo_paper_2024}) that directly aligns the policy's forecasts with actual environmental transitions. Following the success in Deepseek-R1~\cite{deepseek_r1_2025}, we adopt a similar rule-based reward $r_t$ to measures the alignment between the predicted state change $S_\Delta$ and the ground truth $S_\Delta^*$ as.
\begin{equation}
r_t = \frac{\sum_{(k,v_k) \in S_\Delta} w_k \cdot \mathrm{Score}(v_k, v_k^*)}{\sum_{(k,v) \in S_\Delta} w_k}
\end{equation}
where $w_k$ denotes the weight assigned to each key $k$, reflecting its relative importance. The scoring function assigns a value of 1 for exact matches, 0.5 for partial matches, and 0 otherwise. This reward encourages the model to generate action predictions that closely match real player behavior while penalizing overly verbose or irrelevant outputs. We do not incorporate format rewards, as our learned model already demonstrates strong structural adherence. Detailed formulation is provided in Appendix~\ref{appx:rl_grpo}, and training prompts can be found in the Appendix~\ref{sec:training_template}.

\subsection{Inference-Time Decision Making}
\label{sec:inference_time_decision_making}

To leverage the learned world model for strategic gameplay, we implement a \textbf{lookahead search mechanism} during inference (see Figure~\ref{fig:downstream_task_adaption}). Unlike reactive agents that map states directly to actions ($\pi: S \to a$), our approach explicitly reasons about future outcomes before committing to a decision. Specifically, the agent performs a one-step lookahead search:
\begin{itemize}[leftmargin=*]
\setlength\itemsep{0em}
    \item \textbf{Candidate Generation:} Given the current state $S_t$, the model first acts as a policy proposal network, sampling a set of $k$ plausible candidate actions $\mathcal{A}_{cand} = \{a_1, ..., a_k\}$ from the policy distribution $\pi_{\theta}(\cdot | S_t)$.
    \item \textbf{What-if Simulation:} For each candidate action $a_i \in \mathcal{A}_{cand}$, WiA-LLM is queried to forecast the corresponding differential state transition $S_{\Delta, i}$. This step represents the core "What-If" analysis, where the model conditions on the hypothetical execution of $a_i$ to produces $S_{\Delta, i} = f_{\theta}(S_t, a_i)$.
    \item \textbf{Heuristic Evaluation:} To ensure robust selection, we employ a deterministic, rule-based value function $V(S_{\Delta})$. This function acts as a classifier, categorizing the forecasted consequences (e.g., tower destroyed, resource gained, hero death) into strategic values (positive or negative).
    \item \textbf{Optimal Selection:} The final action $a^*$ is selected by maximizing the evaluated outcome of the simulated future: $a^* = \operatorname*{argmax}_{a_i \in \mathcal{A}_{cand}} V(S_{\Delta, i})$. This decouples the generation of possibilities from the evaluation of strategic success, mitigating the risk of the model rationalizing suboptimal plans.
\end{itemize}
This "simulate-and-evaluate" process is the essence of model-based planning. It allows the agent to "think ahead"  and select the action with the best forecasted outcome, rather than simply choosing the action that seems best in a reactive manner. However, we acknowledge that LLM inference introduces a substantial latency, making it impractical to run this process on every game frame. 
To address the latency issue, we employ a dual-system architecture (see Appendix~\ref{sec:latency_analysis}), where WiA-LLM acts as a low-frequency strategic planner (e.g., every 5-10s) guiding a high-frequency reactive policy.

\section{Experiments}
\label{sec:experiment}


\subsection{Experiment Setup}
\label{sec:experiment_setup}

\noindent\textbf{Environment.} All experiments were conducted on four servers, each equipped with 8 NVIDIA H20 GPUs (96 GB each). For SFT, we used the Megatron-LM~\cite{Megatron-LM} training platform, while online RL was performed using OpenRLHF~\cite{hu2024openrlhf}.

\begin{figure*}[t]
     \centering
    \begin{subfigure}[b]{0.48\textwidth}
         \centering
        \includegraphics[width=\textwidth]{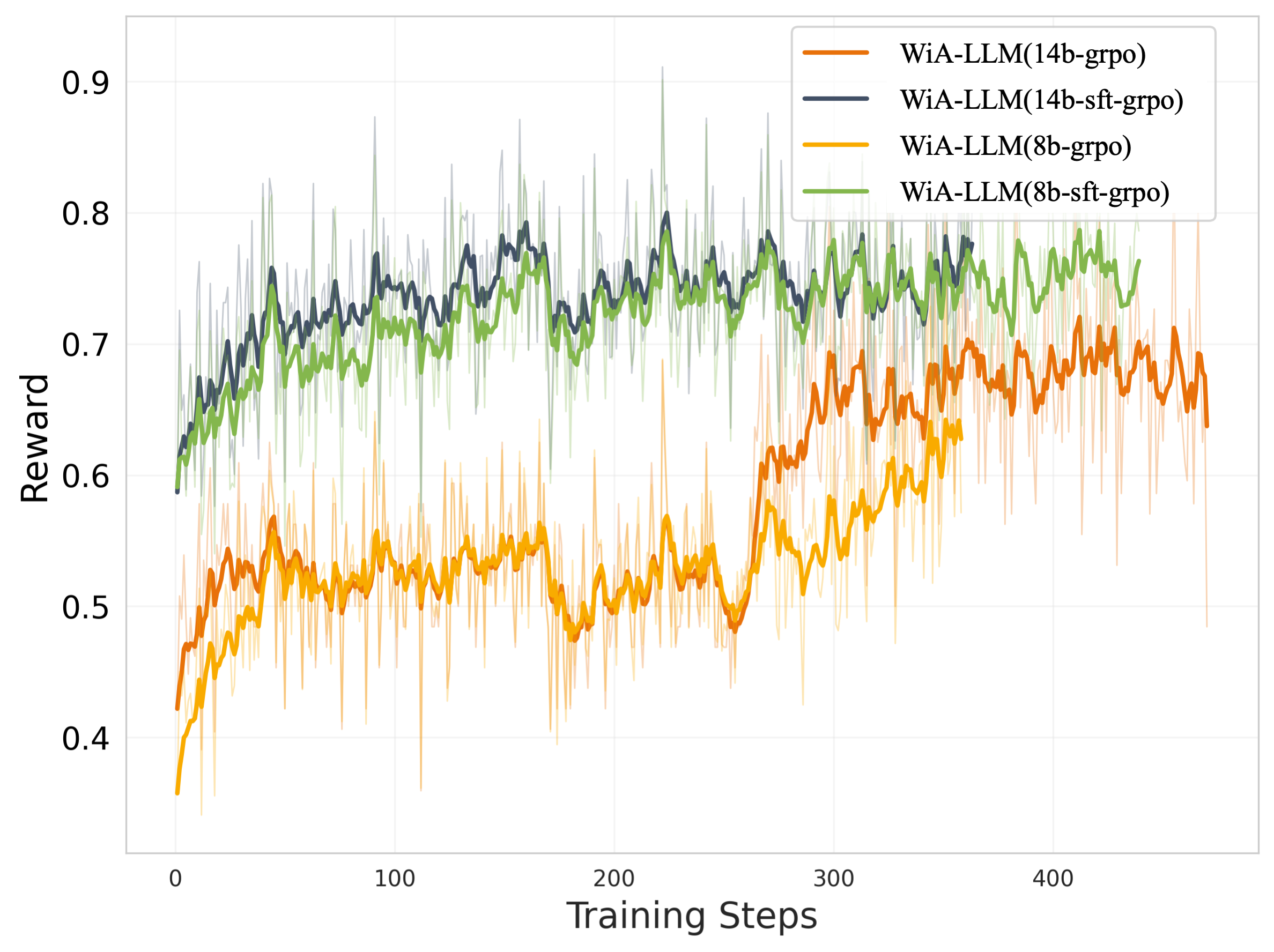}
     \end{subfigure}
     \hfill
     \hfill
     \begin{subfigure}[b]{0.48\textwidth}
         \centering
        \includegraphics[width=\textwidth]{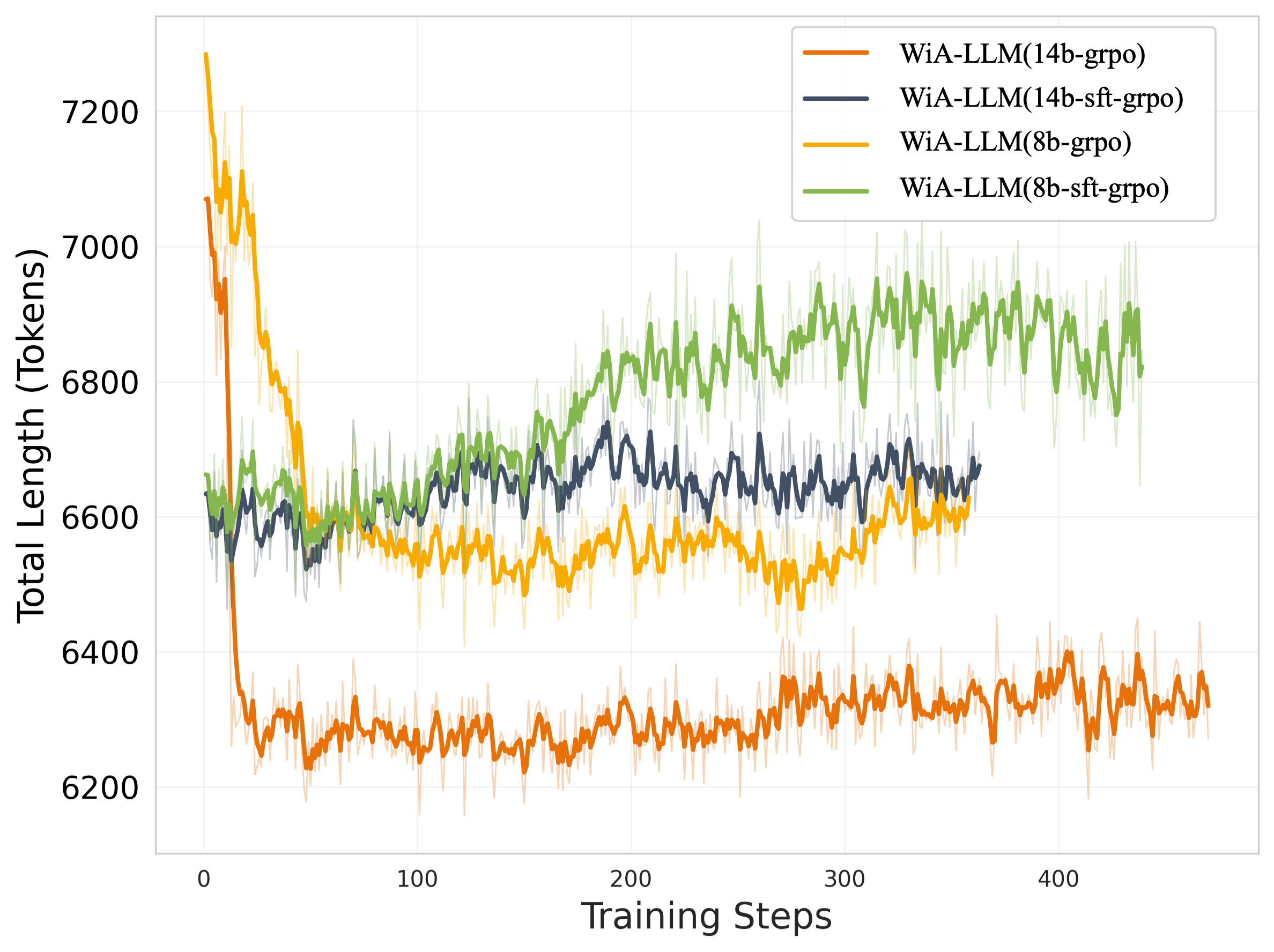}
     \end{subfigure}
     \caption{Demonstration of reward progression (left) and total token length (right) over training steps during the RL process. The results show that our method consistently achieves higher rewards and maintains more stable or longer token lengths compared to the baselines, indicating improved learning efficiency and output quality.}
     \label{fig:rl_training_logs}
\end{figure*}

\noindent\textbf{Datasets.} We curate two new benchmarks for WIA tasks: \textbf{WIA-General} and \textbf{WIA-Hardest}. Task difficulty is quantified by the number of altered game-critical components, denoted as $d = \|S_\Delta\|$, where $S_\Delta$ is the set of differences drawn from $\mathcal{C} = \{\text{hero}, \text{tower}, \text{minion\_waves}, \text{dragon}\}$ as defined in Algorithm~\ref{alg:data_sampling}. The difficulty $d$ ranges from 1 (a single-component change) to 4 (simultaneous changes to all components). The statistics for WIA-General ($\mathcal{D}_g = \{(S_i, a_i, S_\Delta) \mid 1 \leq d \leq 4\}$) and WIA-Hardest ($\mathcal{D}_h = \{(S_i, a_i, S_\Delta) \mid d=4\}$) are provided in Table~\ref{tab:statistics}, with results presented in Table~\ref{tab:performance_wia}. We also verify that domain-specific training does not degrade general capabilities; see Appendix~\ref{sec:general_benchmarks} for details.

\begin{table}[t]
\centering
\resizebox{0.6\linewidth}{!}{
\begin{tabular}{lcccccc}
\toprule
& \multicolumn{2}{c}{\textbf{WIA-General}} & \multicolumn{2}{c}{\textbf{WIA-Hardest}} & \multicolumn{2}{c}{\textbf{Combined}} \\
& \multicolumn{2}{c}{\textbf{(n=1,476)}} & \multicolumn{2}{c}{\textbf{(n=452)}} & \multicolumn{2}{c}{\textbf{(n=1,928)}} \\
\cmidrule(lr){2-3} \cmidrule(lr){4-5} \cmidrule(lr){6-7}
\textbf{Metric} & \textbf{Count} & \textbf{\%} & \textbf{Count} & \textbf{\%} & \textbf{Count} & \textbf{\%} \\
\toprule
\multicolumn{7}{c}{\textbf{Change Types Distribution}} \\
\midrule
Minion Changes & 1,003 & 67.95 & 452 & 100.00 & 1,455 & 75.47 \\
Tower Changes & 1,355 & 91.80 & 452 & 100.00 & 1,807 & 93.73 \\
Hero Changes & 412 & 27.91 & 452 & 100.00 & 864 & 44.81 \\
Dragon Changes & 37 & 2.51 & 452 & 100.00 & 489 & 25.36 \\
\midrule
\multicolumn{7}{c}{\textbf{Difficulty Levels Distribution}} \\
\midrule
d=1 (1 change) & 524 & 35.50 & 0 & 0.00 & 524 & 27.18 \\
d=2 (2 changes) & 567 & 38.41 & 0 & 0.00 & 567 & 29.42 \\
d=3 (3 changes) & 369 & 25.00 & 0 & 0.00 & 369 & 19.15 \\
d=4 (4 changes) & 16 & 1.08 & 452 & 100.00 & 468 & 24.28 \\
\bottomrule
\end{tabular}}
\caption{Statistics: WIA-General \& WIA-Hardest.}
\label{tab:statistics}
\end{table}

\noindent\textbf{Baselines.} We use LLMs of various scales as baselines, primarily focusing on the Qwen3 models~\cite{yang2025qwen3} with native reasoning capabilities, including Qwen3-14B and Qwen3-8B. We also consider several other models: DS-R1, DS-R1-distilled-Qwen3-14B, and QwQ-32B. All model checkpoints are accessible via HuggingFace.

\subsection{Training Details} 
\label{sec:training_details}

Building on insights from Deepseek-R1~\cite{deepseek_r1_2025}, we employ a multi-stage training strategy that combines supervised fine-tuning (SFT) and reinforcement learning (RL) to enhance the capabilities of our language models. Specifically, SFT improves the foundational language understanding and reasoning abilities of our models, while online RL enables efficient exploration and selection of the most effective solutions through trial and error.

We select Qwen3-8B and Qwen3-14B as our base models. For the SFT stage, we curate the training dataset by distilling knowledge from Deepseek-R1, which demonstrates strong reasoning capabilities in game environments and can thoroughly analyzes  game states based on its pre-existing knowledge. Specifically, we provide the ground truth of game state changes and the corresponding instruction to the R1 model, allowing it to generate a reasoning process~\cite{cot_23}, that leads to the final answer. The prompt used for this process is provided in Appendix~\ref{sec:training_template}. The distilled data is formatted as (game state: $S_t$, action: $a_t$, and thinking process: $C_t$), and serve as a valuable resource for training smaller models to acquire R1-like deep reasoning skills. For the online RL stage, we use real gameplay data collected as described in Section~\ref{sec:method}. We explore two setups:(1) applying GRPO directly to the base model without any prior SFT, and (2) applying GRPO to a model that has already been fine-tuned. These two setups are compared in Table~\ref{tab:performance_wia}. Due to computational constraints, we limit GRPO training to approximately 400 steps for all models to ensure a fair comparison, and set the number of epochs for the SFT stage to three.

\subsection{Main Results}
\label{sec:main_results}

The main results are presented in Table~\ref{tab:performance_wia}. 
These results reveal several critical insights into the effectiveness of our multi-stage training approach. Our \ours{} with SFT+GRPO consistently achieves superior performance across all evaluation settings, with particularly notable improvements on the challenging subsets (WIA-Hardest). On WIA-General, both \ours{}-14B and \ours{}-8B with SFT+GRPO attain nearly identical performance (0.742), substantially outperforming the much larger model like Deepseek-R1 (0.326). The performance gain is even more pronounced on WIA-Hardest, where our 8B model achieves 0.426 accuracy compared to Deepseek-R1's 0.111, while the 14B model reaches 0.295. We further evaluate performance across difficulty levels on WIA-General. While all models experience degradation as task complexity increases from $d=1$ to $d=4$, our approach maintains the most robust performance. Notably, on the most challenging $d=4$ task, our 8B model achieves 0.430 accuracy (compared to Deepseek-R1's 0.102), and our 14B model reaches 0.312, which remarkably exceeds the 14B baseline's performance at this difficulty level. These results demonstrate that both SFT and GRPO independently deliver substantial performance improvements; however, when combined, they yield even greater gains, surpassing the results of either method alone.

\begin{table*}[t]
    \centering
    \resizebox{0.85\linewidth}{!}{
    \begin{tabular}{l cc cccc}
    \toprule
    \multirow{2}{*}{\textbf{Model}} & \multicolumn{2}{c}{\textbf{Overall Benchmarks}} & \multicolumn{4}{c}{\textbf{Difficulty Breakdown ($d$)}} \\
    \cmidrule(lr){2-3} \cmidrule(lr){4-7}
     & WIA-General & WIA-Hardest & $d=1$ & $d=2$ & $d=3$ & $d=4$ \\
    \midrule
    deepseek-r1 & 0.326 & 0.111 & 0.443 & 0.298 & 0.213 & 0.102 \\
    deepseek-r1-distilled-qwen-3-14B & 0.370 & 0.046 & 0.566 & 0.339 & 0.156 & 0.023 \\
    qwq-32b & 0.366 & 0.037 & 0.467 & 0.361 & 0.246 & 0.016 \\
    \hdashline\\[-8pt]
    qwen-3-14b  & 0.472 & 0.027 & 0.640 & 0.486 & 0.234 & 0.008 \\
    \ours{} (14b-sft)  & 0.614 & 0.281 & 0.777 & 0.590 & 0.433 & 0.297 \\
    \ours{} (14b-grpo) & 0.674 & 0.132 & 0.895 & 0.660 & 0.404 & 0.117 \\
    \ours{} (14b-sft-grpo) {\scriptsize(our best model)}  & \textbf{0.742} & \textbf{0.295} & \textbf{0.939} & \textbf{0.731} & \textbf{0.497} & \textbf{0.312} \\
    \hdashline\\[-8pt]
    qwen-3-8b & 0.450 & 0.022 & 0.601 & 0.457 & 0.245 & 0.023 \\
    \ours{} (8b-sft) & 0.669 & 0.179 & 0.898 & 0.625 & 0.431 & 0.172 \\
    \ours{} (8b-grpo) & 0.619 & 0.108 & 0.831 & 0.601 & 0.367 & 0.094 \\
    \ours{} (8b-sft-grpo) {\scriptsize(our best model)}  & \textbf{0.742} & \textbf{0.426} & \textbf{0.938} & \textbf{0.728} & \textbf{0.500} & \textbf{0.430} \\
    \bottomrule
    \end{tabular}}
    \caption{Combined performance comparison. The left columns show results on the WIA-General and WIA-Hardest benchmarks. The right columns detail model performance across specific difficulty levels ($d=1$ to $d=4$). Our best models (*-sft-grpo) consistently outperform baselines across all metrics.}
    \label{tab:performance_wia}
\end{table*}

\begin{figure}[ht]
\centering
\includegraphics[width=0.5\linewidth]{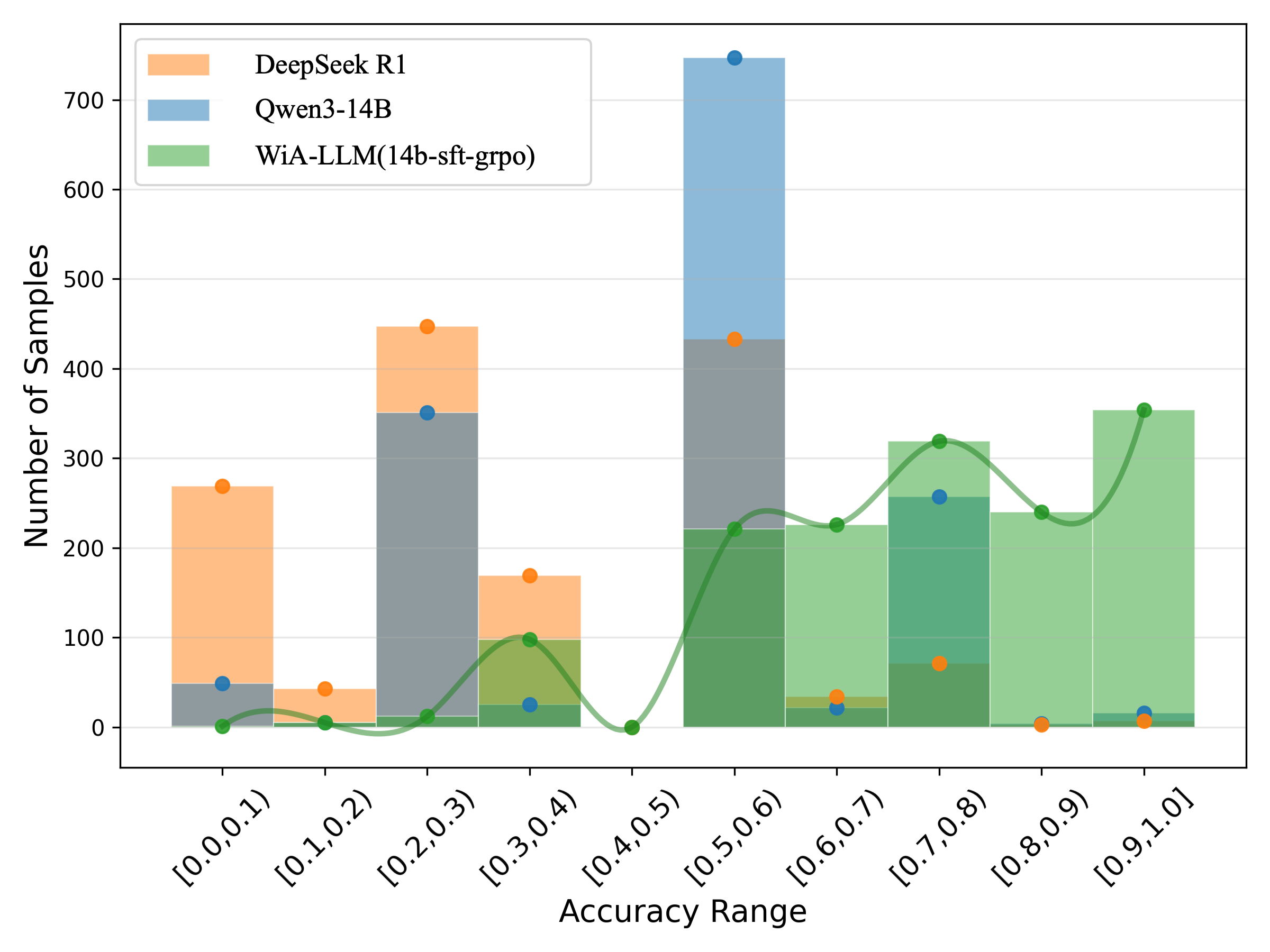}
\caption{Distribution of sample counts across different accuracy ranges. The stacked histograms illustrate the number of samples per accuracy interval, while the smoothed trend line highlights the performance pattern of our method.
}
\label{fig:acc_distribution}
\end{figure}

We also verified that domain-specific training did not degrade general capabilities; see Appendix~\ref{sec:general_benchmarks} for results on MMLU/Math/BBB and other benchmarks.

\subsection{Analysis}
\label{sec:analysis}


\textbf{Response Length vs. Rewards.}
Figure~\ref{fig:rl_training_logs} shows that all models achieve consistent reward improvements, with SFT-initialized variants (\ours{}-*-sft-grpo) starting from significantly higher baselines (0.6 vs. 0.35). While reward trajectories remain largely consistent across models, response lengths differ: SFT models maintain stable or growing output lengths, whereas non-SFT models exhibit initial fluctuations before stabilizing. This suggests that reasoning distillation (SFT) helps models better balance the elaboration of reasoning with reward optimization during RL.

\noindent\textbf{Distribution of Performance Range.}
To further validate the robustness of our model's performance—and to avoid relying solely on aggregate metrics—we examine the distribution of sample counts across different accuracy ranges. As shown in Figure~\ref{fig:acc_distribution}, we plot the sample counts for various models over these ranges and include a smoothed trend line that highlights the performance pattern of \ours{}(14b-sft-grpo). The results demonstrate that our model outperforms baselines by concentrating a larger proportion of samples in the higher accuracy intervals (0.7 to 1.0), indicating more consistent and accurate predictions compared to the base models.

\begin{figure*}[t]
  \centering
  \includegraphics[width=\linewidth]{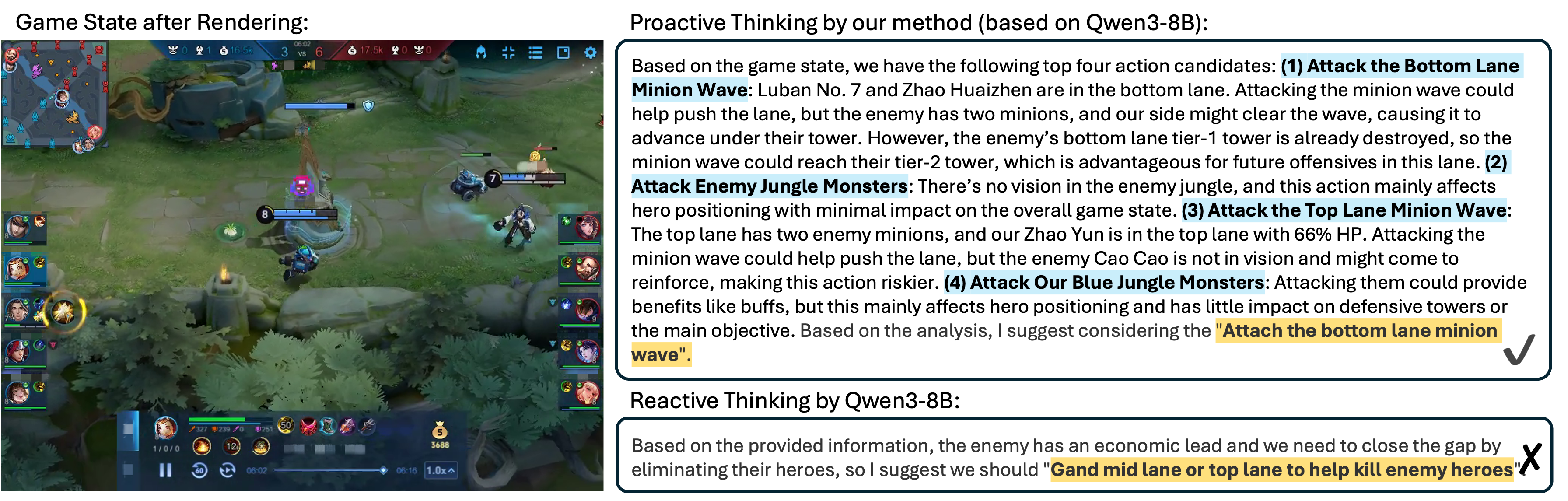}
  \caption{Case study on WiA. To safeguard user privacy, we blurred the user's Game ID.}
  \label{fig:case}
\end{figure*}

\noindent\textbf{Case Studies.}
To assess downstream utility, we conduct a detailed case study illustrated in Figure~\ref{fig:case}. We apply the lookahead search strategy (see Section~\ref{sec:inference_time_decision_making}) to the action prediction task (defined in Section~\ref{sec:definition}). We find that \ours{} produces more detailed and verifiable reasoning traces for each action, and accurately simulates minion wave mechanics to justify a strategic lane push. In contrast, the reactive baseline (Qwen3-8b) hallucinates a non-existent ganking opportunity, underscoring the advantages of proactive reasoning.

\section{Related Works}
\label{sec:related_works}


\noindent\textbf{Game Understanding of LLMs.} 
While LLMs excel at language-based reasoning, applying them effectively to games remains challenging due to their reliance on static pre-training data and lack of environmental grounding~\cite{llm_game_agents_survey_2024}. Key challenges include: (1) \textit{Contextual grounding}—difficulty in interpreting dynamic game states for consistent decision-making~\cite{hu2024pokellmonhumanparityagentpokemon}; (2) \textit{Symbolic precision}—misinterpretation of game terminology and item attributes, which can disrupt interaction with the game engine~\cite{southey2012bayes}; and (3) \textit{Long-term planning}—limited memory and strategic reasoning over extended horizons~\cite{silver2016mastering,starcraft_2017,he2025enabling}. 

\noindent\textbf{Role of RL in LLMs.}
Recent advances in LLMs have highlighted the crucial role of RL in aligning model outputs with human preferences~\cite{meta_reasoner_2025,search_r1_2025}. While pre-training on large text corpora enables LLMs to generate fluent and grammatically correct text, this alone does not guarantee that models are helpful, harmless, or aligned with user expectations. RL from human feedback (RLHF)~\cite{rlhf_2022} addresses this by training a reward model based on human preferences to guide policy optimization via methods such as PPO~\cite{ppo_2017}, DPO~\cite{dpo_2023}, and SimPO~\cite{meng2024simpo}. More recently, GRPO~\cite{grpo_paper_2024} has emerged as a flexible alternative for obtaining reward signals. Unlike PPO, GRPO does not strictly require a reward model; instead, it can incorporate reward signals from any function or model capable of evaluating response quality.

\noindent\textbf{Difference from Time-Series Forecasting.}
Time Series Forecasting (TSF) is a related task that focuses on predicting future values of target variables based on historical data patterns~\cite{tang2025timeseries, shi2022timeseriesforecastingtsf}. Its goal is to extrapolate trends or behaviors from past observations to future timesteps~\cite{Time-LLM}. In contrast, WIA is concerned with understanding the causal impact of specific actions or interventions on the future state of the environment. Rather than simply predicting what will happen next, WIA actively evaluates multiple potential actions to determine which choice leads to the most beneficial outcome. In other words, TSF forecasts the natural progression of a system, while WIA simulates hypothetical scenarios to inform decision-making by assessing the consequences of different actions. This capability is especially valuable for complex decision-making in dynamic environments.




\section{Conclusion \& Limitation}
\label{sec:conclusion}
\textbf{Conclusion:}
We propose \ours{} as an explicit world model that enables LLMs to proactively forecast the consequences of actions. Through interaction with game environments, \ours{} develops a deeper understanding of state dynamics and improves decision-making. While evaluated in Honor of Kings, our approach is broadly applicable to other high-stakes domains where simulating outcomes is safer than trial-and-error.

\textbf{Limitation:}
In this work, we focus on the \textit{strategic reasoning capabilities} of LLMs. While we demonstrate strong alignment with expert human actions and high-fidelity state forecasting, we do not evaluate our approach in an online ranked environment to measure Win Rate. This is because online performance is heavily influenced by low-level execution skills (such as micro-mechanics and reaction time), which fall outside the scope of our study on proactive world modeling.

\clearpage
\bibliography{ref}

@misc{wiki:Partially_observable_Markov_decision_process,
   author = "Wikipedia",
   title = "{Partially observable Markov decision process} --- {W}ikipedia{,} The Free Encyclopedia",
   year = "2025",
   howpublished = {\url{http://en.wikipedia.org/w/index.php?title=Partially\%20observable\%20Markov\%20decision\%20process&oldid=1318763884}},
   note = "[Online; accessed 03-December-2025]"
 }

@misc{grpo_paper_2024,
      title={DeepSeekMath: Pushing the Limits of Mathematical Reasoning in Open Language Models}, 
      author={Zhihong Shao and Peiyi Wang and Qihao Zhu and Runxin Xu and Junxiao Song and Xiao Bi and Haowei Zhang and Mingchuan Zhang and Y. K. Li and Y. Wu and Daya Guo},
      year={2024},
      eprint={2402.03300},
      archivePrefix={arXiv},
      primaryClass={cs.CL},
      url={https://arxiv.org/abs/2402.03300}, 
}

@misc{ape210k,
      title={Ape210K: A Large-Scale and Template-Rich Dataset of Math Word Problems}, 
      author={Wei Zhao and Mingyue Shang and Yang Liu and Liang Wang and Jingming Liu},
      year={2020},
      eprint={2009.11506},
      archivePrefix={arXiv},
      primaryClass={cs.CL},
      url={https://arxiv.org/abs/2009.11506}, 
}

@inproceedings{mmlu,
  author    = {Dan Hendrycks and Collin Burns and Steven Basart and Andy Zou and Mantas Mazeika and Dawn Song and Jacob Steinhardt},
  title     = {Measuring Massive Multitask Language Understanding},
  booktitle = {9th International Conference on Learning Representations, {ICLR} 2021, Virtual Event, Austria, May 3-7, 2021},
  publisher = {OpenReview.net},
  year      = {2021},
  url       = {https://openreview.net/forum?id=d7KBjmI3GmQ},
  bibsource = {dblp computer science bibliography, https://dblp.org}
}

@article{search_r1_2025,
      title   = {Search-R1: Training LLMs to Reason and Leverage Search Engines with Reinforcement Learning},
      author  = {Bowen Jin and Hansi Zeng and Zhenrui Yue and Jinsung Yoon and Sercan Arik and Dong Wang and Hamed Zamani and Jiawei Han},
      year    = {2025},
      journal = {arXiv preprint arXiv: 2503.09516}
}

@misc{meta_reasoner_2025,
      title={Meta-Reasoner: Dynamic Guidance for Optimized Inference-time Reasoning in Large Language Models}, 
      author={Yuan Sui and Yufei He and Tri Cao and Simeng Han and Yulin Chen and Bryan Hooi},
      year={2025},
      eprint={2502.19918},
      archivePrefix={arXiv},
      primaryClass={cs.AI},
      url={https://arxiv.org/abs/2502.19918}, 
}

@inproceedings{ceval,
 author = {Huang, Yuzhen and Bai, Yuzhuo and Zhu, Zhihao and Zhang, Junlei and Zhang, Jinghan and Su, Tangjun and Liu, Junteng and Lv, Chuancheng and Zhang, Yikai and lei, jiayi and Fu, Yao and Sun, Maosong and He, Junxian},
 booktitle = {Advances in Neural Information Processing Systems},
 editor = {A. Oh and T. Naumann and A. Globerson and K. Saenko and M. Hardt and S. Levine},
 pages = {62991--63010},
 publisher = {Curran Associates, Inc.},
 title = {C-Eval: A Multi-Level Multi-Discipline Chinese Evaluation Suite for Foundation Models},
 url = {https://proceedings.neurips.cc/paper_files/paper/2023/file/c6ec1844bec96d6d32ae95ae694e23d8-Paper-Datasets_and_Benchmarks.pdf},
 volume = {36},
 year = {2023}
}

@inproceedings{bbh,
  author    = {Mirac Suzgun and Nathan Scales and Nathanael Sch{\"{a}}rli and Sebastian Gehrmann and Yi Tay and Hyung Won Chung and Aakanksha Chowdhery and Quoc V. Le and Ed H. Chi and Denny Zhou and Jason Wei},
  editor    = {Anna Rogers and Jordan L. Boyd{-}Graber and Naoaki Okazaki},
  title     = {Challenging BIG-Bench Tasks and Whether Chain-of-Thought Can Solve Them},
  booktitle = {Findings of the Association for Computational Linguistics: {ACL} 2023, Toronto, Canada, July 9-14, 2023},
  pages     = {13003-13051},
  publisher = {Association for Computational Linguistics},
  year      = {2023},
  url       = {https://doi.org/10.18653/v1/2023.findings-acl.824},
  doi       = {10.18653/V1/2023.FINDINGS-ACL.824},
  bibsource = {dblp computer science bibliography, https://dblp.org}
}

@article{big-bench,
  author    = {Aarohi Srivastava and Abhinav Rastogi and Abhishek Rao and Abu Awal Md Shoeb and Abubakar Abid and Adam Fisch and Adam R. Brown and Adam Santoro and Aditya Gupta and others},
  title     = {Beyond the Imitation Game: Quantifying and extrapolating the capabilities of language models},
  journal   = {Trans. Mach. Learn. Res.},
  volume    = {2023},
  year      = {2023},
  url       = {https://openreview.net/forum?id=uyTL5Bvosj},
  bibsource = {dblp computer science bibliography, https://dblp.org}
}

@inproceedings{hok_off_2023,
      author    = {Qu, Yun and Wang, Boyuan and Shao, Jianzhun and Jiang, Yuhang and Chen, Chen and Ye, Zhenbin and Linc, Liu and Feng, Yang and Lai, Lin and Qin, Hongyang and Deng, Minwen and Zhuo, Juchao and Ye and others},
      booktitle = {Advances in Neural Information Processing Systems},
      editor    = {A. Oh and T. Naumann and A. Globerson and K. Saenko and M. Hardt and S. Levine},
      pages     = {22166-22190},
      publisher = {Curran Associates, Inc.},
      title     = {Hokoff: Real Game Dataset from Honor of Kings and its Offline Reinforcement Learning Benchmarks},
      url       = {https://proceedings.neurips.cc/paper_files/paper/2023/file/464fefa022aaefc85d901317bbf13f85-Paper-Datasets_and_Benchmarks.pdf},
      volume    = {36},
      year      = {2023}
}

@inproceedings{hok_env_2022,
      author    = {Hua Wei and Jingxiao Chen and Xiyang Ji and Hongyang Qin and Minwen Deng and Siqin Li and Liang Wang and Weinan Zhang and Yong Yu and Liu Lin and others},
      editor    = {Sanmi Koyejo and S. Mohamed and A. Agarwal and Danielle Belgrave and K. Cho and A. Oh},
      title     = {Honor of Kings Arena: an Environment for Generalization in Competitive Reinforcement Learning},
      booktitle = {Advances in Neural Information Processing Systems 35: Annual Conference on Neural Information Processing Systems 2022, NeurIPS 2022, New Orleans, LA, USA, November 28 - December 9, 2022},
      year      = {2022},
      url       = {http://papers.nips.cc/paper\_files/paper/2022/hash/4dbb61cb68671edc4ca3712d70083b9f-Abstract-Datasets\_and\_Benchmarks.html},
      bibsource = {dblp computer science bibliography, https://dblp.org}
    }

@article{li2025codei0o0,
      title   = {CodeI/O: Condensing Reasoning Patterns via Code Input-Output Prediction},
      author  = {Junlong Li and Daya Guo and Dejian Yang and Runxin Xu and Yu Wu and Junxian He},
      year    = {2025},
      journal = {arXiv preprint arXiv: 2502.07316}
}

@article{charactereval,
  title   = {CharacterEval: A Chinese Benchmark for Role-Playing Conversational Agent Evaluation},
  author  = {Quan Tu and Shilong Fan and Zihang Tian and Rui Yan},
  year    = {2024},
  journal = {arXiv preprint arXiv: 2401.01275}
}

@article{ifeval,
      title   = {Instruction-Following Evaluation for Large Language Models},
      author  = {Jeffrey Zhou and Tianjian Lu and Swaroop Mishra and Siddhartha Brahma and Sujoy Basu and Yi Luan and Denny Zhou and Le Hou},
      year    = {2023},
      journal = {arXiv preprint arXiv: 2311.07911}
}

@inproceedings{rlhf_2022,
      author    = {Long Ouyang and Jeffrey Wu and Xu Jiang and Diogo Almeida and Carroll L. Wainwright and Pamela Mishkin and Chong Zhang and Sandhini Agarwal and Katarina Slama and others},
      editor    = {Sanmi Koyejo and S. Mohamed and A. Agarwal and Danielle Belgrave and K. Cho and A. Oh},
      title     = {Training language models to follow instructions with human feedback},
      booktitle = {Advances in Neural Information Processing Systems 35: Annual Conference on Neural Information Processing Systems 2022, NeurIPS 2022, New Orleans, LA, USA, November 28 - December 9, 2022},
      year      = {2022},
      url       = {http://papers.nips.cc/paper\_files/paper/2022/hash/b1efde53be364a73914f58805a001731-Abstract-Conference.html},
      bibsource = {dblp computer science bibliography, https://dblp.org}
}

@inproceedings{dpo_2023,
      author    = {Rafael Rafailov and Archit Sharma and Eric Mitchell and Christopher D. Manning and Stefano Ermon and Chelsea Finn},
      editor    = {Alice Oh and Tristan Naumann and Amir Globerson and Kate Saenko and Moritz Hardt and Sergey Levine},
      title     = {Direct Preference Optimization: Your Language Model is Secretly a Reward Model},
      booktitle = {Advances in Neural Information Processing Systems 36: Annual Conference on Neural Information Processing Systems 2023, NeurIPS 2023, New Orleans, LA, USA, December 10 - 16, 2023},
      year      = {2023},
      url       = {http://papers.nips.cc/paper\_files/paper/2023/hash/a85b405ed65c6477a4fe8302b5e06ce7-Abstract-Conference.html},
      bibsource = {dblp computer science bibliography, https://dblp.org}
}

@article{ppo_2017,
      title   = {Proximal Policy Optimization Algorithms},
      author  = {John Schulman and Filip Wolski and Prafulla Dhariwal and Alec Radford and Oleg Klimov},
      year    = {2017},
      journal = {arXiv preprint arXiv: 1707.06347}
}

@article{meng2024simpo,
      title   = {Simpo: Simple preference optimization with a reference-free reward},
      author  = {Meng, Yu and Xia, Mengzhou and Chen, Danqi},
      journal = {Advances in Neural Information Processing Systems},
      volume  = {37},
      pages   = {124198-124235},
      year    = {2024}
}

@article{llm_game_agents_survey_2024,
      title   = {A Survey on Large Language Model-Based Game Agents},
      author  = {Sihao Hu and Tiansheng Huang and Gaowen Liu and Ramana Rao Kompella and Fatih Ilhan and Selim Furkan Tekin and Yichang Xu and Zachary Yahn and Ling Liu},
      year    = {2024},
      journal = {arXiv preprint arXiv: 2404.02039}
}

@misc{cot_23,
      title={Chain-of-Thought Prompting Elicits Reasoning in Large Language Models}, 
      author={Jason Wei and Xuezhi Wang and Dale Schuurmans and Maarten Bosma and Brian Ichter and Fei Xia and Ed Chi and Quoc Le and Denny Zhou},
      year={2023},
      eprint={2201.11903},
      archivePrefix={arXiv},
      primaryClass={cs.CL},
      url={https://arxiv.org/abs/2201.11903}, 
}

@Online{school_chinese,
  author   = {{lanhin}},
  title    = {School Chinese Benchmark},
  url      = {https://github.com/lanhin/SchoolChinese},
  year = {2018}
}

@article{Megatron-LM,
  title   = {Megatron-LM: Training Multi-Billion Parameter Language Models Using Model Parallelism},
  author  = {Mohammad Shoeybi and Mostofa Patwary and Raul Puri and Patrick LeGresley and Jared Casper and Bryan Catanzaro},
  year    = {2019},
  journal = {arXiv preprint arXiv: 1909.08053}
}

@article{hu2024openrlhf,
  title={OpenRLHF: An Easy-to-use, Scalable and High-performance RLHF Framework},
  author={Jian Hu and Xibin Wu and Zilin Zhu and Xianyu and Weixun Wang and Dehao Zhang and Yu Cao},
  journal={arXiv preprint arXiv:2405.11143},
  year={2024}
}

@misc{starcraft_2017,
      title={StarCraft II: A New Challenge for Reinforcement Learning}, 
      author={Oriol Vinyals and Timo Ewalds and Sergey Bartunov and Petko Georgiev and others},
      year={2017},
      eprint={1708.04782},
      archivePrefix={arXiv},
      primaryClass={cs.LG},
      url={https://arxiv.org/abs/1708.04782}, 
}

@article{silver2016mastering,
      title={Mastering the game of Go with deep neural networks and tree search},
      author={Silver, David and Huang, Aja and Maddison, Chris J and Guez, Arthur and Sifre, Laurent and Van Den Driessche, George and Schrittwieser, Julian and Antonoglou, Ioannis and Panneershelvam, Veda and Lanctot, Marc and others},
      journal={nature},
      volume={529},
      number={7587},
      pages={484--489},
      year={2016},
      publisher={Nature Publishing Group}
}

@article{southey2012bayes,
      title={Bayes' bluff: Opponent modelling in poker},
      author={Southey, Finnegan and Bowling, Michael P and Larson, Bryce and Piccione, Carmelo and Burch, Neil and Billings, Darse and Rayner, Chris},
      journal={arXiv preprint arXiv:1207.1411},
      year={2012}
}

@misc{deepseek_r1_2025,
      title={DeepSeek-R1: Incentivizing Reasoning Capability in LLMs via Reinforcement Learning}, 
      author={DeepSeek-AI and Daya Guo and Dejian Yang and Haowei Zhang and Junxiao Song and Ruoyu Zhang and Runxin Xu and others},
      year={2025},
      eprint={2501.12948},
      archivePrefix={arXiv},
      primaryClass={cs.CL},
      url={https://arxiv.org/abs/2501.12948}, 
}

@misc{hu2024pokellmonhumanparityagentpokemon,
      title={PokeLLMon: A Human-Parity Agent for Pokemon Battles with Large Language Models}, 
      author={Sihao Hu and Tiansheng Huang and Ling Liu},
      year={2024},
      eprint={2402.01118},
      archivePrefix={arXiv},
      primaryClass={cs.AI},
      url={https://arxiv.org/abs/2402.01118}, 
}

@misc{what-if-analysis-business,
      title={Augmenting Decision Making via Interactive What-If Analysis}, 
      author={Sneha Gathani and Madelon Hulsebos and James Gale and Peter J. Haas and Çağatay Demiralp},
      year={2022},
      eprint={2109.06160},
      archivePrefix={arXiv},
      primaryClass={cs.DB},
      url={https://arxiv.org/abs/2109.06160}, 
}

@article{tang2025timeseries,
  title={Time series forecasting with llms: Understanding and enhancing model capabilities},
  author={Tang, Hua and Zhang, Chong and Jin, Mingyu and Yu, Qinkai and Wang, Zhenting and Jin, Xiaobo and Zhang, Yongfeng and Du, Mengnan},
  journal={ACM SIGKDD Explorations Newsletter},
  volume={26},
  number={2},
  pages={109--118},
  year={2025},
  publisher={ACM New York, NY, USA}
}

@misc{shi2022timeseriesforecastingtsf,
      title={Time Series Forecasting (TSF) Using Various Deep Learning Models}, 
      author={Jimeng Shi and Mahek Jain and Giri Narasimhan},
      year={2022},
      eprint={2204.11115},
      archivePrefix={arXiv},
      primaryClass={cs.LG},
      url={https://arxiv.org/abs/2204.11115}, 
}

@inproceedings{Time-LLM,
  author    = {Ming Jin and Shiyu Wang and Lintao Ma and Zhixuan Chu and James Y. Zhang and Xiaoming Shi and Pin{-}Yu Chen and Yuxuan Liang and Yuan{-}Fang Li and Shirui Pan and Qingsong Wen},
  title     = {Time-LLM: Time Series Forecasting by Reprogramming Large Language Models},
  booktitle = {The Twelfth International Conference on Learning Representations, {ICLR} 2024, Vienna, Austria, May 7-11, 2024},
  publisher = {OpenReview.net},
  year      = {2024},
  url       = {https://openreview.net/forum?id=Unb5CVPtae},
  bibsource = {dblp computer science bibliography, https://dblp.org}
}

@article{liao2025think,
  title   = {Think in Games: Learning to Reason in Games via Reinforcement Learning with Large Language Models},
  author  = {Yi Liao and Yu Gu and Yuan Sui and Zining Zhu and Yifan Lu and Guohua Tang and Zhongqian Sun and Wei Yang},
  year    = {2025},
  journal = {arXiv preprint arXiv: 2508.21365}
}

@article{yang2025qwen3,
  title   = {Qwen3 Technical Report},
  author  = {An Yang and Anfeng Li and Baosong Yang and Beichen Zhang and Binyuan Hui and Bo Zheng and Bowen Yu and Chang Gao and Chengen Huang and Chenxu Lv and Chujie Zheng and Dayiheng Liu and Fan Zhou and Fei Huang and Feng Hu and Hao Ge and Haoran Wei and Huan Lin and Jialong Tang and Jian Yang and Jianhong Tu and Jianwei Zhang and Jianxin Yang and Jiaxi Yang and Jing Zhou and Jingren Zhou and Junyang Lin and Kai Dang and Keqin Bao and Kexin Yang and Le Yu and Lianghao Deng and Mei Li and Mingfeng Xue and Mingze Li and Pei Zhang and Peng Wang and Qin Zhu and Rui Men and Ruize Gao and Shixuan Liu and Shuang Luo and Tianhao Li and Tianyi Tang and Wenbiao Yin and Xingzhang Ren and Xinyu Wang and Xinyu Zhang and Xuancheng Ren and Yang Fan and Yang Su and Yichang Zhang and Yinger Zhang and Yu Wan and Yuqiong Liu and Zekun Wang and Zeyu Cui and Zhenru Zhang and Zhipeng Zhou and Zihan Qiu},
  year    = {2025},
  journal = {arXiv preprint arXiv: 2505.09388}
}

@article{he2025enabling,
  title   = {Enabling Self-Improving Agents to Learn at Test Time With Human-In-The-Loop Guidance},
  author  = {Yufei He and Ruoyu Li and Alex Chen and Yue Liu and Yulin Chen and Yuan Sui and Cheng Chen and Yi Zhu and Luca Luo and Frank Yang and Bryan Hooi},
  year    = {2025},
  journal = {arXiv preprint arXiv: 2507.17131}
}

\appendix
\clearpage

\section{Definition}
\label{sec:definition}

In this section, we formally define the What-If Analysis task within the game environment and introduce the key notations used in our framework.

\subsection{Task Definition: What-If Analysis}
Given a game state $S_t$ at time step $t$, the objective is to forecast the resulting state change, denoted as $S_\Delta$, after the player executes a specific action $a_t$. This task requires the model to reason about the current environmental conditions and demonstrate a deep understanding of game mechanics to predict the causal impact of each action. As illustrated in Figure~\ref{fig:demonstration}, WIA reflects a shift toward \textit{proactive thinking}: instead of simply reacting to the current state, the model anticipates the future consequences of its decisions, thus enabling more informed strategic planning.

\subsection{Task Definition: Action Prediction}
\label{sec:downstream_task}
We consider action prediction as one of the downstream tasks in game decision-making. This task can be formalized as $a_t^* = f(S_t)$, where the model predicts the optimal action based on the current game state $S_t$. In the HoK environment, the available actions are defined as a set $A$ (see Table~\ref{tab:action_space} for details). Notably, these actions are designed at the strategic level, providing high-level guidance to users who then adjust their low-level operations (such as moving the hero or using abilities) accordingly. This work differs from prior studies~\cite{hok_env_2022,hok_off_2023}, which primarily focus on low-level actions like changing hero positions and skill releases.

\subsection{Game State}
We model the game environment as a sequence of discrete states. Each state $S_t$ encapsulates all visible information from the player's perspective, including teammate attributes, visible turrets, and map vision. To ensure realistic gameplay, we strictly enforce \textbf{partial observability} by excluding hidden information, such as the status of enemies concealed by the "fog of war." For compatibility with LLMs, the game state $S_t$ is serialized into a JSON object—a format that LLMs can naturally parse and process.

\subsection{Game Environment}
We conduct our study in \textit{Honor of Kings (HoK)}~\cite{hok_env_2022, hok_off_2023}. In HoK, players control unique heroes and coordinate with teammates to defeat opponents, neutral creatures, and defensive structures, with the ultimate goal of destroying the opposing team's base crystal. HoK serves as an ideal testbed due to its complexity: the need for team coordination, dynamic strategic shifts, and the high-dimensional state space present significant challenges for decision-making. Mastering proactive reasoning in such a complex domain holds great promise for advancing the reasoning capabilities of game AI.

\begin{algorithm}[t]
\caption{Game State Parsing and Difference Extraction}
\label{alg:data_sampling}
\begin{algorithmic}[1]
\small
\Require Raw Game State Sequence $S = [S_1, \dots, S_T]$
\Ensure Transition Dataset $G_D$
\State \textbf{Constant} $\mathcal{C} \leftarrow \{\text{hero}, \text{tower}, \text{minions}, \text{dragon}\}$
\State Initialize lists: $P \leftarrow [\ ]$; $G_D \leftarrow [\ ]$
\For{$t \leftarrow 1$ \textbf{to} $T$} \Comment{\textit{Phase 1: Annotate Actions}}
    \State $a_t \leftarrow \Call{Annotate}{S_t}$ 
    \State $P.\Call{Append}{(S_t, a_t)}$
\EndFor
\For{$i \leftarrow 1$ \textbf{to} $|P|-1$} \Comment{\textit{Phase 2: Extract Transitions}}
    \State Let $(S_{cur}, a_{cur}) = P[i]$ and $(S_{next}, a_{next}) = P[i+1]$
    
    \If{$\Call{TimeDelta}{P[i], P[i+1]} > 60\text{s}$}
        \State \textbf{continue} 
    \EndIf

    \If{$a_{cur} \neq a_{next}$}
        \State $S_\Delta \leftarrow \Call{ComputeStateDiff}{S_{cur}, S_{next}}$
        \State $G_D.\Call{Append}{(S_{cur}, a_{cur}, S_\Delta)}$
    \EndIf
\EndFor
\State \Return $G_D$
\Function{ComputeStateDiff}{$S_{\text{old}}, S_{\text{new}}$}
    \State $S_\Delta \leftarrow \text{Map}()$
    \For{\textbf{each} $c \in \mathcal{C}$}
        \If{$S_{\text{old}}[c] \neq S_{\text{new}}[c]$}
            \State $S_\Delta[c] \leftarrow \Call{Diff}{S_{\text{old}}[c], S_{\text{new}}[c]}$
        \EndIf
    \EndFor
    \State \Return $S_\Delta$
\EndFunction
\end{algorithmic}
\end{algorithm}

\begin{figure*}[t]
  \centering
  \includegraphics[width=0.9\linewidth]{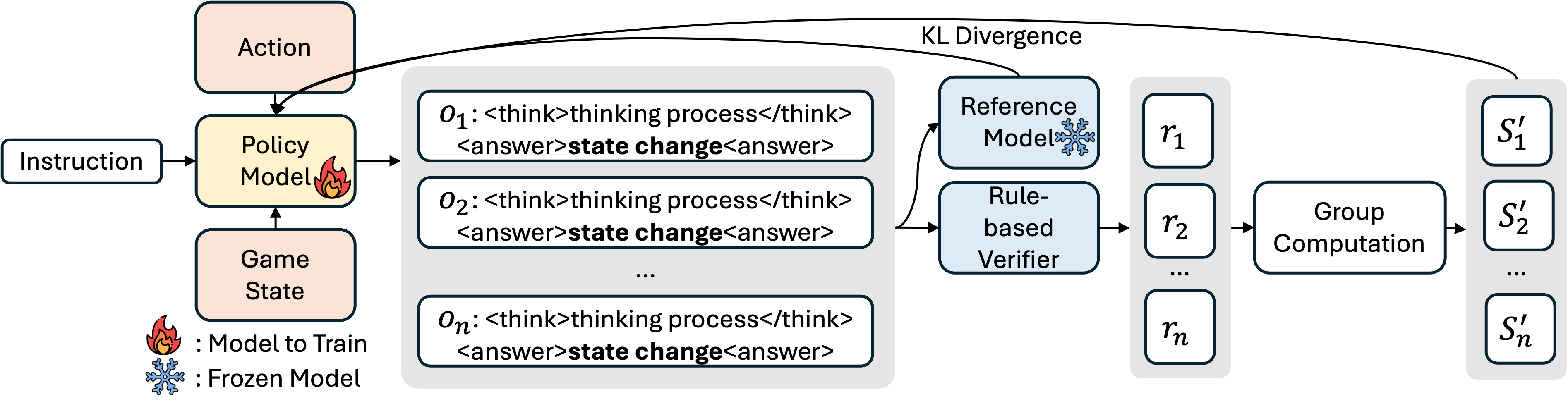}
  \caption{\textbf{Demonstration} of GRPO training with Game State. 
    Given the player's action and the current game state, the model is asked to forecast the potential changes to the entire game state once the player takes the action, and provide the thinking process as the analysis of this what-if scenario. We then use the predicted game state changes to compare with ground-truth values using a rule-based verifier to update the policy model. This process enables the model to perform what-if analysis (forecasting) by simulating action outcomes and refining its decision-making accordingly.
  }
  \label{fig:grpo}
\end{figure*}

\begin{figure*}[t]
    \centering
    \includegraphics[width=0.85\linewidth]{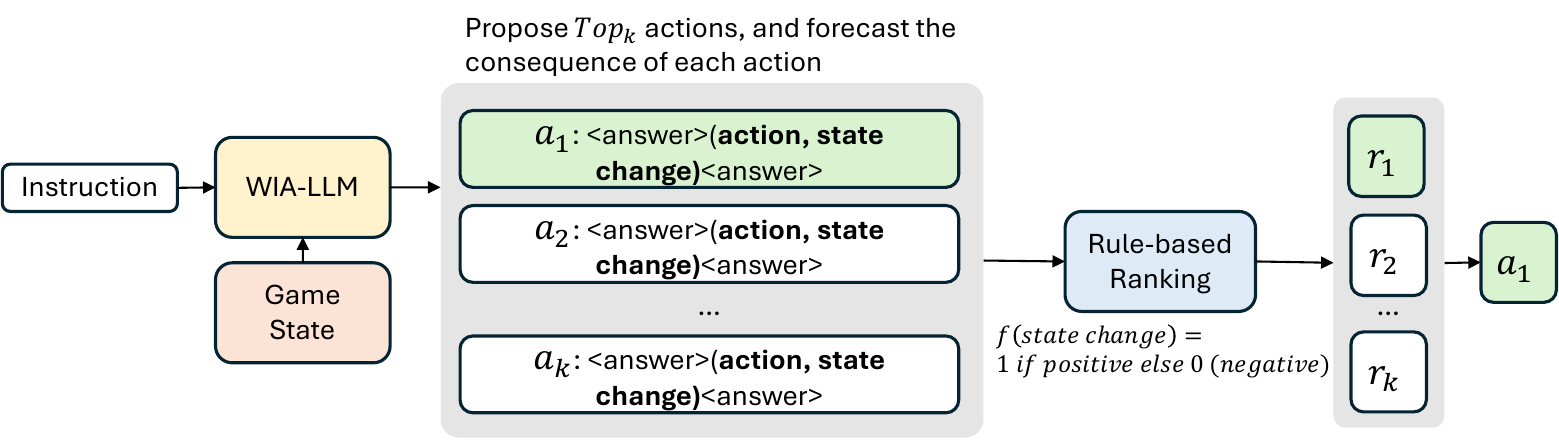}
    \caption{\textbf{Demonstration}: Adaption of \ours{} to downstream action prediction task.}
    \label{fig:downstream_task_adaption}
\end{figure*}

\section{GRPO Formulation with Game State}
\label{appx:rl_grpo}

To facilitate effective learning of proactive reasoning in game environments, we employ \emph{Group Relative Policy Optimization} (GRPO)~\cite{grpo_paper_2024}, an online RL algorithm designed to maximize the advantage of generated completions while constraining policy divergence from a reference model.

We formalize the training process of \ours{} using GRPO as follows. Let $q$ denote a sampled prompt (e.g., a game state $s_t$ and context $i_t$), and let $\{o_1, o_2, \dots, o_G\}$ represent a group of $G$ completions generated by the old policy $\pi_{\theta_{\text{old}}}$. For each completion $o_i$, a reward $r_i$ is computed using a rule-based reward function (see Section~\ref{sec:rl_mtd}). The group-relative advantage for each completion is then calculated as:
\begin{equation}
\hat{A}_{i,t} = \frac{r_i - \text{mean}(r)}{\text{std}(r)},
\end{equation}
where $\text{mean}(r)$ and $\text{std}(r)$ denote the mean and standard deviation of rewards within the group, respectively. This normalization ensures the advantage reflects the relative quality of each completion.

To optimize the policy, we first define the importance sampling ratio $\rho_t(\theta)$ between the current policy $\pi_\theta$ and the sampling policy $\pi_{\theta_{\text{old}}}$:
\begin{equation}
\label{eq:ratio_def}
\rho_t(\theta) = \frac{\pi_\theta(o_{i,t} | q, o_{i,<t})}{\pi_{\theta_{\text{old}}}(o_{i,t} | q, o_{i,<t})}.
\end{equation}

Additionally, to ensure the policy remains close to the original behavior, we compute the token-level Kullback-Leibler (KL) divergence between the current policy $\pi_\theta$ and the reference policy $\pi_{\text{ref}}$:
\begin{equation}
\label{eq:kl_def}
\small
\mathbb{D}_{\text{KL}} \left[ \pi_\theta \| \pi_{\text{ref}} \right] = \pi_\theta(o_{i,t} | q, o_{i,<t}) \log \frac{\pi_\theta(o_{i,t} | q, o_{i,<t})}{\pi_{\text{ref}}(o_{i,t} | q, o_{i,<t})}.
\end{equation}
Alternatively, this can be approximated using the estimator $\frac{\pi_{\text{ref}}}{\pi_\theta} - \log \frac{\pi_{\text{ref}}}{\pi_\theta} - 1$~\cite{ppo_2017}.

The overall GRPO objective is to maximize the expected group-relative advantage while penalizing deviations from the reference model. We define the total objective $\mathcal{J}_{\text{GRPO}}$ as:
\begin{equation}
\label{eq:grpo_objective}
    \mathcal{J}_{\text{GRPO}}(\theta) = \hat{\mathbb{E}}_{i,t} \left[ \mathcal{L}^{\text{CLIP}}_t(\theta) - \beta \mathbb{D}_{\mathrm{KL}}[\pi_\theta \| \pi_{\text{ref}}] \right]
\end{equation}
where $\beta$ is a coefficient controlling the strength of the KL regularization. To ensure training stability, we employ the clipped surrogate objective $\mathcal{L}^{\text{CLIP}}_t(\theta)$, defined as:
\begin{equation}
\label{eq:clip_def}
    \min \left( \rho_t(\theta) \hat{A}_{i,t}, \, \text{clip}(\rho_t(\theta), 1-\epsilon, 1+\epsilon) \hat{A}_{i,t} \right)
\end{equation}
Here, the clipping operator constrains the update magnitude relative to the old policy within the range $[1-\epsilon, 1+\epsilon]$, thereby preventing destructive policy updates.

\section{Latency \& Real-Time Inference}
\label{sec:latency_analysis}
A core strength of \ours{} is its ability to perform high-fidelity, interpretable counterfactual reasoning. However, as discussed in \refsec{sec:inference_time_decision_making}, LLM inference introduces significant latency, making it impractical to run on every game frame. To address this, we propose a two-pronged strategy: (1) a dual-system deployment architecture, and (2) knowledge distillation for scalable deployment.

\subsection{Dual-System Deployment Architecture}
We do not propose running the full \ours{} simulation for low-level, high-frequency actions (e.g., movement or skill targeting). Instead, the agent operates under a dual-system architecture that separates slow, deliberative reasoning from fast, reactive execution.

\noindent\textbf{System 1 (Reactive Policy):} A lightweight, model-free neural network handles real-time, low-latency control of the hero (e.g., movement and immediate tactical skill use). This policy can operate at the required environment frequency (e.g., $\sim 30$ ms per update).

\noindent\textbf{System 2 (WiA-LLM Planner):} WiA-LLM acts as the deliberative strategic layer. The planner is invoked only at key decision points that require long-horizon reasoning. Typical calls include: (i) \textbf{objective selection}, such as deciding whether to contest Dragon/Baron, push a high-ground tower, or rotate lanes; (ii) \textbf{rallying and grouping}, such as deciding whether to commit to a team fight; and (iii) \textbf{recall and resurrection planning}, such as choosing the hero’s path and next objective after death or returning to base. This design reduces LLM inference frequency from every frame to roughly once every 5--10 seconds, or at major state transitions (e.g., an enemy hero death). In this way, the agent benefits from strategic planning without sacrificing real-time control.

\subsection{Knowledge Distillation from \ours{}}
The primary value of WiA-LLM is its ability to internalize complex game dynamics and predict strategic outcomes with high precision, though at a computational cost. To enable scalable deployment, we leverage knowledge distillation to transfer this strategic understanding into a much smaller, faster student model. Specifically, WiA-LLM (the teacher model) is used to generate a large, high-quality dataset of reasoning-augmented trajectories, including the current state, optimal high-level actions (from what-if analysis), and textual justifications. A much smaller, efficient neural network (the student model) is then trained to mimic the teacher’s strategic decisions. 

\textbf{However}, while we present this as the natural path to production deployment, the detailed implementation of knowledge distillation is beyond the scope of this work. Here, we focus on demonstrating the core capability of WiA-LLM to learn and execute explicit world modeling, leaving knowledge distillation for future work.

\begin{table}[t]
    \centering
    \resizebox{0.6\linewidth}{!}{
    \begin{tabular}{ll}
    \toprule
        \textbf{Dataset}  &  \textbf{URL} \\
    \midrule
        \textbf{Ape210K}~\cite{ape210k}  & https://github.com/Chenny0808/ape210k\\
        \textbf{MMLU}~\cite{mmlu} & https://huggingface.co/datasets/cais/mmlu\\
        \textbf{CEval}~\cite{ceval} & https://github.com/hkust-nlp/ceval\\
        \textbf{School-Chinese}~\cite{school_chinese} & https://github.com/lanhin/SchoolChinese\\
        \textbf{BBH}~\cite{bbh} & https://github.com/suzgunmirac/BIG-Bench-Hard\\
        \textbf{IfEval}~\cite{ifeval}  & https://huggingface.co/datasets/google/IFEval\\
        \textbf{CharacterEval}~\cite{charactereval} & https://github.com/morecry/charactereval\\
    \bottomrule
    \end{tabular}}
    \caption{Source Link for the Benchmarks}
    \label{tab:dataset_urls}
\end{table}

\section{\ours{} on General Benchmarks}
\label{sec:general_benchmarks}
To further verify that our models do not sacrifice their native language understanding and reasoning capabilities during training, we evaluate \ours{} on several standard benchmarks: \textbf{Ape210K}~\cite{ape210k}, \textbf{MMLU}~\cite{mmlu}, \textbf{CEval}~\cite{ceval}, \textbf{School-Chinese}~\cite{school_chinese}, \textbf{BBH}~\cite{bbh}, and \textbf{IfEval}~\cite{ifeval}.

\begin{table}[ht]
\centering
\resizebox{0.6\linewidth}{!}{
    \begin{tabular}{lcccc}
    \toprule
    \multirow{2}{*}{\textbf{Model}} & \textbf{\thead{Prompt-level\\loose-acc}} & \textbf{\thead{Inst-level\\loose-acc}} & \textbf{\thead{Prompt-level\\strict-acc}} & \textbf{\thead{Inst-level\\strict-acc}} \\
    \midrule
    qwen3-14b & 0.357 & 0.494 & 0.338 & 0.475 \\
    \ours{} (14b-grpo) & 0.355 & 0.496 & 0.342 & 0.482  \\
    \ours{} (14b-sft) & 0.355 & 0.498 & 0.336 & 0.480 \\
    \ours{} (14b-sft-grpo)  & 0.362 & 0.501 & 0.344 & 0.486 \\
    \hdashline\\[-8pt]
    qwen3-8b & 0.351 & 0.495 & 0.340 & 0.483 \\	
    \ours{} (8b-grpo) & 0.379 & 0.514 & 0.366 & 0.500 \\	
    \ours{} (8b-sft) & 0.301 & 0.440 & 0.290 & 0.429 \\
    \ours{} (8b-sft-grpo) & 0.336 &  0.477 & 0.323 & 0.461 \\		
    \bottomrule
    \end{tabular}}
\caption{Performance on the IFEval benchmark~\cite{ifeval}. Detailed metrics are provided in Appendix~\ref{sec:ifeval_metrics}.}
\label{tab:perf_ifeval_benchmark}
\end{table}

\subsection{Performance Analysis}
Tables~\ref{tab:perf_academic_knowledge_benchmark} and \ref{tab:perf_ifeval_benchmark} show that our training approach generally preserves core language model capabilities, sometimes even enhancing logical reasoning and instruction-following. On math tasks (Ape210K), our 14B models remain stable (93.5 across all variants), while the 8B SFT-based models show only modest degradation. Subject exams (MMLU and CEval) demonstrate exceptional robustness, with performance variations within 1 percentage point, indicating factual knowledge is well preserved. Notably, logical reasoning (BBH) consistently improves with SFT-based training, with 8B models achieving 60.17–60.52 compared to the 58.35 baseline. In Table~\ref{tab:perf_ifeval_benchmark}, we analyze instruction-following capabilities, finding that our method with GRPO enhances performance (8B: 0.379 vs. 0.351 on prompt-level accuracy), while SFT alone may cause slight degradation on the 8B model (0.301 vs. 0.351). These results confirm that our approach enables domain-specific improvements while preserving essential language model abilities, with RL components proving particularly beneficial for maintaining instruction-following performance.

\begin{table}[ht]
\centering
\resizebox{0.6\linewidth}{!}{
    \begin{tabular}{lccccc}
    \toprule
    \multirow{2}{*}{\textbf{Model}} & \textbf{Math} & \textbf{Memorization} & \multicolumn{2}{c}{\textbf{Subject Exam}} & \textbf{Logic} \\
    \cmidrule(lr){2-2}  \cmidrule(lr){3-3} \cmidrule(lr){4-5} \cmidrule(lr){6-6}
    & Ape\_210k & SchoolChinese & MMLU & CEval & BBH \\
    \midrule
    qwen3-14b & \textbf{93.5} & 91.85 & 80.21 & 82.76 & \textbf{65.48} \\
    \ours{} (14b-sft) & 92.0 & \textbf{92.28} & 80.18 & 83.06 & \textbf{65.48} \\ 
    \ours{} (14b-grpo) & \textbf{93.5} & 92.19 & \textbf{80.56} & 82.91 & 65.13 \\ 
    \ours{} (14b-sft-grpo) & \textbf{93.5} & 91.69 & 80.25 & \textbf{83.14} & 65.30 \\
    \hdashline\\[-8pt]
    qwen3-8b &  93.0 & 88.0 & 75.96 & 78.08 & 58.35 \\
    \ours{} (8b-sft) & 90.5 & 87.02 & 76.60 & \textbf{78.68} & \textbf{60.52} \\ 
    \ours{} (8b-grpo) & \textbf{93.5} & \textbf{88.04} & 76.04 & 78.45 & 57.83 \\ 
    \ours{} (8b-sft-grpo) & 89.5 & 86.72 & \textbf{76.631} & 78.6 & 60.17 \\ 
    \bottomrule
    \end{tabular}
}
\caption{Performance of different models on math, academic, general knowledge, and logical reasoning benchmarks.}
\label{tab:perf_academic_knowledge_benchmark}
\end{table}

\subsection{Benchmarks Descriptions}
\label{sec:benchmarks}
The details of the benchmarks are as follows. To facilitate reproducibility, we provide the source links for these benchmarks in Table~\ref{tab:dataset_urls}.
\begin{itemize}[leftmargin=*]
\item \textbf{Ape210K}~\cite{ape210k}: A large-scale, template-rich math word problem dataset. For our experiments, we randomly sample 200 examples from the test set.
\item \textbf{MMLU}~\cite{mmlu}: A comprehensive benchmark covering knowledge from 57 subjects across STEM, humanities, social sciences, and more. It ranges in difficulty from elementary to advanced professional level, testing both world knowledge and problem-solving ability. We sample the first 50 examples from each subject, resulting in $50 \times 57 = 2850$ cases for our experiments.
\item \textbf{CEval}~\cite{ceval}: Similar to MMLU, CEval is a Chinese-language benchmark comprising 52 subtasks across four categories: STEM, social sciences, humanities, and others. We use it as an additional testbed to evaluate language mixing challenges as discussed in~\cite{deepseek_r1_2025}.
\item \textbf{School-Chinese}~\cite{school_chinese}: This benchmark assesses the memorization capabilities of LLMs on classical Chinese poetry by requiring the model to predict subsequent content given introductory text. We manually collected these datasets from public repositories, resulting in a benchmark with 269 samples.
\item \textbf{BBH}~\cite{bbh}: A subset of BIG-Bench~\cite{big-bench} focused on 23 challenging tasks that require multi-step reasoning. It is widely regarded as a standard evaluation set for assessing the logical reasoning abilities of language models.
\item \textbf{IfEval}~\cite{ifeval}: A standard benchmark for evaluating the instruction-following capabilities of LLMs. It contains approximately 500 verifiable instructions, such as "write more than 400 words" or "mention the keyword 'AI' at least three times," which can be automatically checked using heuristics.
\end{itemize}

\subsection{Metrics for IFEval}
\label{sec:ifeval_metrics}
We use the IFEval benchmark~\cite{ifeval} to evaluate the instruction-following capabilities of \ours{}, as reported in Table~\ref{tab:perf_ifeval_benchmark}. The evaluation is based on four key metrics, which vary along two dimensions: \textbf{granularity} (whether evaluation is performed at the level of the entire prompt or on individual instructions) and \textbf{verification strictness} (how rigorously the model's output is checked).

For granularity, we consider two levels: (1) \textit{Prompt-Level}: The entire prompt (which may contain multiple instructions) is evaluated as a single unit. All instructions must be satisfied for the prompt to be counted as correct. (2) \textit{Instruction-Level}: Each instruction is evaluated independently, regardless of the prompt to which it belongs.

For verification strictness, we also consider two levels: (1) \textit{Strict}: The output must match the instructions exactly, including both content and formatting. Even minor deviations (e.g., missing bold text) result in failure. (2) \textit{Loose}: Some flexibility is allowed by applying transformations to the output (e.g., removing markdown or extra text) prior to evaluation, focusing on the main intent of the instruction.

The precise definitions of each metric are as follows:
\begin{itemize}[leftmargin=*]
\item \textbf{Prompt-Level Strict-Accuracy}: The percentage of prompts for which every instruction is followed exactly as specified, with no deviations in content or formatting. For prompts with multiple instructions, all must be perfectly executed for the prompt to be counted as correct; a single error causes the prompt to fail.
\item \textbf{Instruction-Level Strict-Accuracy}: The percentage of individual instructions across all prompts that are followed exactly as specified. Each instruction is checked independently, and the metric counts how many are perfectly executed, even if others in the same prompt fail.
\item \textbf{Prompt-Level Loose-Accuracy}: A more lenient version of prompt-level accuracy. After normalizing the output (e.g., removing markdown or extraneous text), all instructions must be satisfied for the entire prompt to be counted as correct.
\item \textbf{Instruction-Level Loose-Accuracy}: A more relaxed version of instruction-level accuracy. After normalization, each instruction is evaluated separately and considered correct if it meets the requirements, even if there are minor formatting differences.
\end{itemize}


\newtcolorbox{InlinePromptBox}[2][]{
  enhanced,
  breakable,
  colback=ysshallowblue,  
  colframe=ysdarkblue,    
  title={#2},
  fonttitle=\bfseries,
  attach boxed title to top left={yshift=-2mm, xshift=5mm},
  boxed title style={boxrule=0pt, colframe=white, colback=ysdarkblue, sharp corners},
  sharp corners=south,
  arc=3mm,
  drop lifted shadow,
  #1
}

\newtcolorbox{ActionTableBox}[2][]{
  enhanced,
  breakable,
  colback=white,          
  colframe=ysdarkred,     
  colbacktitle=ysdarkred,
  title={#2},
  fonttitle=\bfseries\Large,
  sharp corners,
  boxrule=1.5pt,
  #1
}


\section{Prompting List}
\label{sec:training_template}

\paragraph{Training Prompt.}
To train \ours{}, we begin by designing a straightforward template that guides the initial LLM to follow our predefined instructions. As shown in Table~\ref{box:training_prompt}, this template organizes the model's output into two parts: a reasoning process and a final answer. We intentionally restrict our requirements to this structural format, following~\cite{search_r1_2025}, in order to avoid introducing any content-specific biases.

\begin{InlinePromptBox}[label={box:training_prompt}]{Training Prompt}
\begin{Verbatim}
You are an AI assistant for the Honor of Kings game. As the main player's assistant, you need to analyze the potential battlefield changes #time_gap# seconds after the main player's action. 

The current game state is: <game_state>GAME_STATE</game_state>, and the main player's executed action is <action>ACTION</action>.

Please consider the following four aspects when analyzing the potential game state change:
  - 1. Minion Wave Changes: Analyze changes in minion wave pushing status (e.g., whether a lane's minions enter or exit enemy turret range) and changes in wave-clearing heroes.
  - 2. Turret Changes: Analyze changes in turret HP and changes in turret attack status. Note that turret protection mechanics do not grant invincibility; turrets still lose HP when attacked.
  - 3. Hero Changes: Analyze hero deaths (e.g., a hero being eliminated).
  - 4. Dragon Status Changes: Analyze whether the Lord, Turtle, or Storm Dragon is being attacked. Note that any HP reduction is considered an attack.

Please put your thinking process in <think></think>, and game state change in <answer></answer>. The answer should be formulated as a JSON object covering the following keys: (1) minion_wave_changes; (2) turret_changes; (3) hero_changes and (4) dragon_status_changes.
\end{Verbatim}
\end{InlinePromptBox}

\vspace{1em}

\paragraph{Distillation Prompt from DS-R1.}
In Table~\ref{box:Distillation_Prompt}, we present the prompt used to distill reasoning data from Deepseek-R1. We supply the ground-truth game state change along with the corresponding instructions to the R1 model, prompting it to generate the reasoning process ($C_t$) that links the state ($S_t$) and action ($a_t$) to the resulting outcome.

\begin{InlinePromptBox}[label={box:Distillation_Prompt}]{Distillation Prompt from DS-R1}
\begin{Verbatim}
You are an AI assistant for the Honor of Kings game. Your task is to analyze the given battlefield changes and generate the corresponding logical reasoning process explaining how these changes occurred based on the current game state. 

Current game state: <game_state>GAME_STATE</game_state>; Main play's executed action: <action>ACTION</action>;

Game state changes after #time_gap#: <game_state_change>STATE_CHANGE</game_state_change>. 

As a game assistant, you need to analyze the causes of the battlefield changes that occurred #time_gap# seconds after the main player's action and generate a reasoning process explaining how these changes happened based on the current game state. 

Place your reasoning inside <answer></answer>.
\end{Verbatim}
\end{InlinePromptBox}

\vspace{1em}

\paragraph{Downstream Task Prompt.}
In Table~\ref{tab:prompt_template_downstream_task}, we present the prompt used for the downstream task of action prediction. The model first selects the top four potential actions and then analyzes the consequences of each to determine the optimal choice.

\begin{InlinePromptBox}[label={tab:prompt_template_downstream_task}]{Downstream Task Prompt}
\begin{Verbatim}
Given real-time game state information from an MOBA game, provide decision-making suggestions as an assistant to the main player.
    
# Board State: <game_state>GAME_STATE</game_state>

Please firstly select 4 most probable actions from the candidate options set <action_candidates>ACTION_CANDIDATES</action_candidates>. 

And then analyse the consequences of each action one by one. 

In the last, select the action most beneficial to our team based on whether the consequences are favorable to our situation.  

Please put your thinking process in <think></think>, and actions in <answer></answer>.
\end{Verbatim}
\end{InlinePromptBox}

\clearpage

\begin{ActionTableBox}[label={tab:action_space}]{Action Space for Downstream Tasks}
\renewcommand{\arraystretch}{1.2}
\begin{tabularx}{\linewidth}{c l X}
\toprule
\textbf{Category} & \textbf{Action} & \textbf{Explanation} \\ 
\midrule
\textbf{None} & None & No action triggered for a short period \\ 
\midrule
\multirow{3}{*}{\textbf{Dragon}} & Lord & Deal damage to the Lord (Main Dragon) \\ 
 & Tyrant & Deal damage to the Tyrant (Early Game Dragon) \\ 
 & Dragon King & Deal damage to the Dragon King (Late Game Dragon) \\ 
\midrule
\multirow{4}{*}{\textbf{Tower}} & Crystal & Deal damage to enemy Crystal (Nexus) \\ 
 & Top Tower & Deal damage to Top Lane Tower \\ 
 & Mid Tower & Deal damage to Mid Lane Tower \\ 
 & Bot Tower & Deal damage to Bottom Lane Tower \\ 
\midrule
\multirow{4}{*}{\textbf{Defense}} & Defend Crystal & Defend our Crystal \\ 
 & Defend Top Tower & Defend Top Lane Tower \\ 
 & Defend Mid Tower & Defend Mid Lane Tower \\ 
 & Defend Bot Tower & Defend Bottom Lane Tower \\ 
\midrule
\multirow{5}{*}{\textbf{Hero}} & Top/Mid/Bot Hero & Damage enemy heroes in respective lanes \\ 
 & River Top/Bot Hero & Damage enemies in River areas \\ 
 & Allied/Enemy Jungle Hero & Damage enemies in Jungle areas \\ 
 & Ally High-ground Hero & Damage enemies on our High-ground \\ 
 & Enemy High-ground Hero & Damage enemies on enemy High-ground \\ 
\midrule
\multirow{3}{*}{\textbf{Line}} & Top/Mid/Bot Minions & Clear minions in respective lanes \\ 
 & Ally High-ground Minions & Clear minions on our High-ground \\ 
 & Enemy High-ground Minions & Clear minions on enemy High-ground \\ 
\midrule
\multirow{2}{*}{\textbf{Buff}} & Allied Red/Blue & Take our Red/Blue Buff \\ 
 & Enemy Red/Blue & Steal enemy Red/Blue Buff \\ 
\midrule
\multirow{2}{*}{\textbf{Jungle}} & Allied/Enemy Camps & Clear non-buff camps \\ 
 & Void Spirit / Crimson Raptor & Kill River objectives (Crabs) \\ 
\midrule
\multirow{4}{*}{\textbf{Grouping}} & Lane Grouping & Group in Top, Mid, or Bot Lane \\ 
 & River Grouping & Group in Upper or Lower River \\ 
 & Jungle Grouping & Group in Allied or Enemy Jungle \\ 
 & High-ground Grouping & Group on Allied or Enemy High-ground \\ 
\midrule
\textbf{Recall} & Recall & Hero at fountain (including walk-back) \\ 
\bottomrule
\end{tabularx}
\end{ActionTableBox}

\end{document}